\documentclass{article}
\usepackage[english]{babel}
\usepackage[utf8]{inputenc}
\usepackage{johd}
\usepackage{listings}
\usepackage{subcaption}

\hypersetup{
    colorlinks,
    linktoc=none,
    linkcolor=blue,
    citecolor=blue,
    urlcolor=blue
}

\newcommand*{\footnotemarkcolor}{black}
\makeatletter
\renewcommand*{\@makefnmark}{\hbox{\@textsuperscript{%
   \color{\footnotemarkcolor}\normalfont\@thefnmark}}}
\makeatother

\title{Enhancing Large Language Model Performance To Answer Questions and Extract Information More Accurately}

\author{Liang Zhang, Katherine Jijo, Spurthi Setty, Eden Chung, \\ Fatima Javid, Natan Vidra, Tommy Clifford}

\date{} 

\begin{document}

\maketitle

\begin{abstract} 
\noindent 
Large Language Models (LLMs) generate responses to questions; however, their effectiveness is often hindered by sub-optimal quality of answers and occasional failures to provide accurate responses to questions. To address these challenges, a fine-tuning process is employed, involving feedback and examples to refine models. The objective is to enhance AI models through continuous feedback loops, utilizing metrics such as cosine similarity, LLM evaluation and Rouge-L scores to evaluate the models. Leveraging LLMs like GPT-3.5, GPT4ALL, and LLaMA2, and Claude, this approach is benchmarked on financial datasets, including the FinanceBench and RAG Instruct Benchmark Tester Dataset, illustrating the necessity of fine-tuning. The results showcase the capability of fine-tuned models to surpass the accuracy of zero-shot LLMs, providing superior question and answering capabilities. Notably, the combination of fine-tuning the LLM with a process known as Retrieval Augmented Generation (RAG) proves to generate responses with improved accuracy.

\end{abstract}


\section{Introduction}

\noindent In the world of language models like ChatGPT, a significant challenge is figuring out if these models' responses are correct or if the models are hallucinating—providing random or incorrect answers. Large Language Models (LLMs) represent a significant leap in deep learning, characterized by their expansive size and pre-training on diverse datasets. Built on the transformer architecture, LLMs such as GPT-3 and ChatGPT, possess encoder and decoder components with self-attention capabilities, enabling them to extract meanings and comprehend relationships within text sequences. The unsupervised training of transformer LLMs facilitates self-learning, encompassing basic grammar, languages, and knowledge. The scalability of the transformer architecture accommodates very large models, often containing billions of parameters, capable of ingesting extensive datasets from diverse sources. The versatility of LLMs is showcased in their ability to perform tasks ranging from question answering to code generation. Once LLMs complete their initial training, they possess the remarkable capability of adapting to various tasks through a process known as fine-tuning. Fine-tuning involves adjusting the model's parameters using additional, more specific data, allowing it to specialize in a particular application. For example, with a singular LLM such as GPT-3, it can be fine-tuned for various use cases, such as answering questions in different industries such as finance or healthcare, or answering questions in simple terms for elementary school-level students. There are three common learning models that showcase the versatility of LLMs: zero-shot learning, where the base models can respond broadly to various requests without specific training; few-shot learning, where providing a small set of relevant examples significantly improves performance in a specific area; and fine-tuning, which extends few-shot learning by allowing data scientists to train the base model with additional data tailored to the specific application, refining its parameters for optimal performance in diverse tasks. This adaptability makes LLMs powerful tools that can be fine-tuned to excel in specific domains and meet varied requirements effectively.
\\

\noindent The current concern with the employment of LLMs is the notion of hallucinations and their implications. LLMs tend to distort information or offer inaccurate answers. In real-world applications, the discrepancies in the information that a model provides due to hallucinations will have significant consequences. Consider a situation where a financial analyst consults a large language model regarding a company's fiscal data, particularly inquiring about the reported revenue. The correct answer is "2.37M," reflecting 2.37 million dollars. However, due to hallucination, the model might answer with an altered value, such as "23.7M." A subtle shift in a decimal point, due to hallucination, drastically distorts the representation of the reported revenue, presenting an exaggerated and false figure. \\
\\
\noindent In the realm of financial decision-making, where numerical accuracy is paramount - relying on inaccurate data to drive decisions will result in misguided investment choices, financial planning errors, or misjudgments of a company's performance. It is imperative to resolve hallucination through the application of methods like fine-tuning to allow language models to provide reliable information in domains where the stakes are high and accuracy is crucial.

\section{Background}
To understand different techniques to enhance large language model performance to extract information and answer questions more accurately, we looked into information about RAG, Fine Tuning, PEFT, LORA, QLORA and other methods.

\subsection{Retrieval Augmented Generation}
 Retrieval Augmented generation (RAG) emerges as a crucial process in optimizing the output of large language models. LLMs, with their vast training data and billions of parameters, excel at tasks like question answering, language translation, and sentence completion. However, inherent challenges include the generation of inaccurate or outdated responses, presenting false information, and a lack of adaptability to current events. RAG addresses these issues by extending the capabilities of LLMs to reference authoritative knowledge bases outside their training data, enhancing relevance, accuracy, and usefulness in various contexts. RAG takes an input, retrieves a set of relevant/supporting documents from a source like for example your textbook pdf, and combines them with the original input prompt. This concatenated context is then fed to the text generator, producing the final output. This adaptability of RAG becomes valuable in situations where facts may change over time, a feature particularly useful as the parametric knowledge of LLMs remains static. RAG eliminates the need for retraining, allowing language models to access the latest information for generating reliable outputs through retrieval-based generation. introduced in ``Retrieval-Augmented Generation for Knowledge-Intensive NLP Tasks " \footnote{Patrick Lewis, Ethan Perez, Aleksandra Piktus, Fabio Petroni, Vladimir Karpukhin, Naman Goyal, Heinrich Küttler, Mike Lewis, Wen-tau Yih, Tim Rocktäschel, Sebastian Riedel, Douwe Kiela. Retrieval-Augmented Generation for Knowledge-Intensive NLP Tasks, 2020.}
\\

\noindent The RAG model faces many limitations that impact its effectiveness in knowledge-intensive natural language processing tasks. Semantic search, a core component of RAG, exhibits challenges such as retrieving irrelevant or opposing information, indicating the model's sensitivity to language nuances and the potential for unexpected results. The ambiguity in understanding how the embedding model extracts and organizes information in vectors adds complexity to optimizing similarity functions. Additionally, the process of chunking, crucial in RAG, can result in information loss if not carefully designed. \footnote{Disadvantages of RAG. \url{https://medium.com/@kelvin.lu.au/disadvantages-of-rag-5024692f2c53}, Accessed on 2024-01-24.}
\\

\begin{figure}[h!]
\centering
\caption{Retrieval-Augmented Generation\protect\footnotemark}
\includegraphics[scale=0.45]{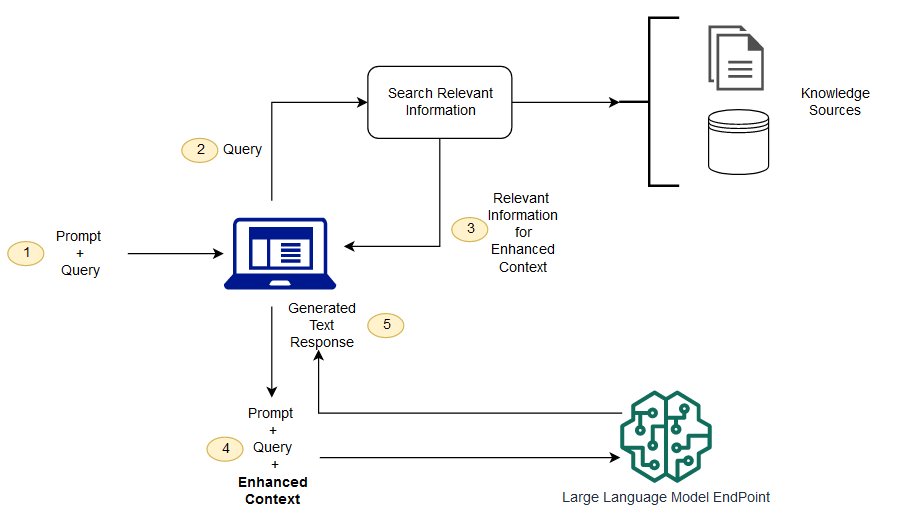}
\end{figure} 

\footnotetext{Retrieval Augmented Generation (RAG). \url{https://docs.aws.amazon.com/sagemaker/latest/dg/jumpstart-foundation-models-customize-rag.html}}


\subsection{Fine Tuning}

While Retrieval Augmented Generation provides a method of injecting new knowledge sources into the LLM, the model itself remains unchanged which is why RAG alone has subpar results on very domain specific data. LLMs like GPT or LLaMA are not trained with a specific purpose in mind, yet many users would like to use LLMs for a specific purpose; in our case, we wanted to achieve  domain-specific question answering from documents. Fine-tuning enables us to customize the output of the model based on specific use cases without having to go through the entire training process again. Actual fine-tuning alters the weights of the model based on the additional fine-tuning data, while RAG edits the query without making any changes to the model. So, a combination of these two methods could optimize the model for a specific use-case. There are various methods for fine-tuning large language models, such as Supervised Fine-tuning (SFT) via Parameter Efficient Fine Tuning (PEFT) \\

\noindent Supervised Fine-tuning is when the training data consists of paired input and outputs of various examples that demonstrate how you want the model to respond to certain queries. By fine-tuning the model, we aim to get the model to mimic the style of the training data when asked questions similar to those in the data. Since our goal for the fine-tuned model was to be good at answering questions on financial documents, our training data consisted of a set of questions and answers curated by human financial experts. The datasets we employed were meant for supervised fine-tuning as they consisted of a question, context, and answer column. 
\\

\noindent Parameter Efficient Fine Tuning is a method that significantly reduces the amount of compute and memory required for Fine Tuning LLMs. Instead of retraining and adjusting all the weights, PEFT freezes all the weights of the pre-trained model and then augments it with additional parameters during the fine tuning process. This differs from full fine tuning which retrains all the parameter weights, and transfer learning which only retrains the head of the model. PEFT can result in comparable performance to fully fine tuned models while having significantly less trainable parameters. It also reduces the risk of a model forgetting a lot of core material and enables the same base model to be fine tuned for various different use cases since the all the base pre-trained weights remain fixed. \footnote{ Sourab Mangrulkar and Sayak Paul, “Parameter-Efficient Fine-Tuning Using PEFT,” Parameter-Efficient Fine-Tuning using PEFT, February 10, 2023, \url {https://huggingface.co/blog/peft.}} \\

\noindent Low Rank Adaptation (LoRA) is a method within PEFT that further reduces the number of trainable parameters through the use of low-rank matrices. Instead of representing the weights in a d x d matrix for example, LoRA represents it as the product of (d x r)(r x d), where r is a rank significantly lower than d. Compared to full fine tuning, LoRA uses 10000 times less trainable parameters and three times less memory. \footnote{Edward Hu, Yelong Shen, Phillip Wallis, Zeyuan Allen-Zhu, Yuanzhi Li Shean Wang Lu Wang Weizhu Chen, LORA: LOW-RANK ADAPTATION OF LARGE LANGUAGE MODELS, 2021} PEFT with LoRA is significantly less computationally expensive that full fine tuning and has been proven to result in comparable performance, as seen in Figure 2. Therefore, such techniques have been widely adapted and the ability to fine tune LLMs for specific use cases has been dramatically democratized. \\

\noindent QLoRA, or Quantized Low-Rank Adaptation is a method that further increases the efficiency of fine tuning by significantly reducing the memory required while preserving performance through the use of quantization. Tensors in these models are usually in 16-bit precision and a quantized model then reduces the precision and memory of these models, usually to 4-bit precision. QLoRA quantizes each input tensor of the neural network and backpropagates gradients through a this quantized model into Low Rank Adapters. \footnote{Tim Dettmers, Artidoro Pagnoni,Ari Holtzman
Luke Zettlemoyer, QLORA: Efficient Finetuning of Quantized LLMs, 2023} 
\\

\begin{figure}[h!]
\centering
\caption{Fine Tuning Methods and LoRA performance \protect\footnotemark}
\includegraphics[scale=0.80]{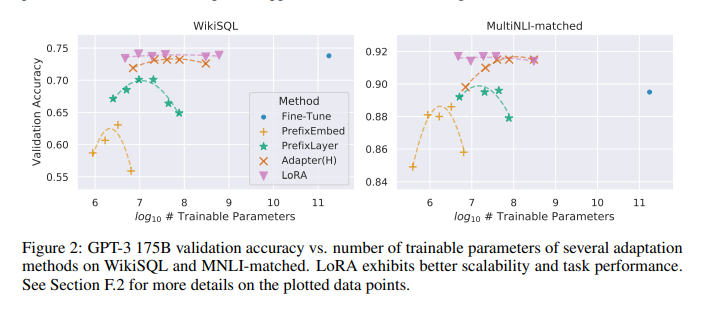}
\end{figure}

\begin{figure}[h!]
\centering
\caption{Parameter Efficient Fine Tuning \protect\footnotemark}
\includegraphics[scale=0.60]{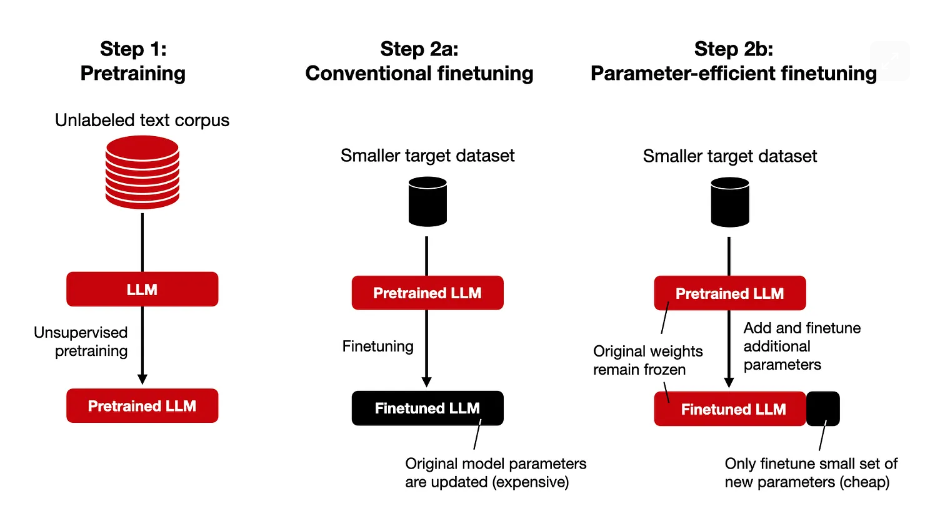}
\end{figure}

\section{Process and Work}

To enhance the accuracy of question and answer (Q\&A) tasks, our approach included supervised fine-tuning on the zero-shot model for GPT-3.5 Turbo. This is needed because zero shot Retrieval-Augmented Generation (RAG) oftentimes retrieves irrelevant information or irrelevant embeddings. Subsequently, we utilized reprompting on GPT4All and Llama2 for further refinement. This process included feeding in example prompts with given questions, evidence text, and answers to guide the model in learning specific response patterns. Additionally, we benchmarked the fine-tuning process by incorporating the Retrieval-Augmented Generation (RAG) technique for Q\&A tasks, thereby enhancing the evaluation and overall performance of the models.
\\

\noindent There is an absence in standardized methodologies for evaluating model performance. We assessed the zero-shot model (LLM without fine-tuning) and subsequently evaluated the model with fine-tuning and reprompting. The approach involved iterative accuracy evaluation, incorporating fine-tuning or reprompting into models such as GPT-3.5 Turbo, GPT4All, Llama2, and Claude. The evaluation metrics used were cosine similarity and Rouge-L, measuring accuracy in each iteration

\subsection{Datasets}

For fine-tuning and testing, we utilized two question-answering datasets. FinanceBench is a novel benchmark developed by Patronus AI for evaluating the performance of LLMs on open-book financial question answering. The dataset is comprised of 10,231 questions about publicly traded companies. Each question is accompanied by corresponding answers and evidence strings. It covers 40 companies in the USA, spanning 361 public filings, including 10Ks, 10Qs, 8Ks, and Earnings Reports, released between 2015 and 2023. A typical entry contains the question (e.g., ``What Was AMCOR's Adjusted Non-GAAP EBITDA for FY 2023"), the answer (e.g., ``AMCOR's Adj. EBITDA was 2,018 million USD in FY 2023"), an evidence string (containing information needed to verify the answer), and a page number from the relevant document. The second dataset is RAG Instruct Benchmark tester, designed by LLMWARE to measure the different capabilities of retrieval augmented generation in financial and legal enterprise use cases. The dataset includes 200 questions with context passages pulled from common 'retrieval scenarios', e.g., financial news, earnings releases, contracts, invoices, technical articles, general news and short texts. The questions, spanning Core Q\&A Evaluation, Not Found Classification, Boolean - Yes/No, Math, Complex Q\&A and Summary categories, are strategically segmented to assess the capabilities of LLMs.

\begin{figure}[h!]
\centering
\caption{Sample of FinanceBench Dataset \protect\footnotemark}
\includegraphics[scale=0.45]{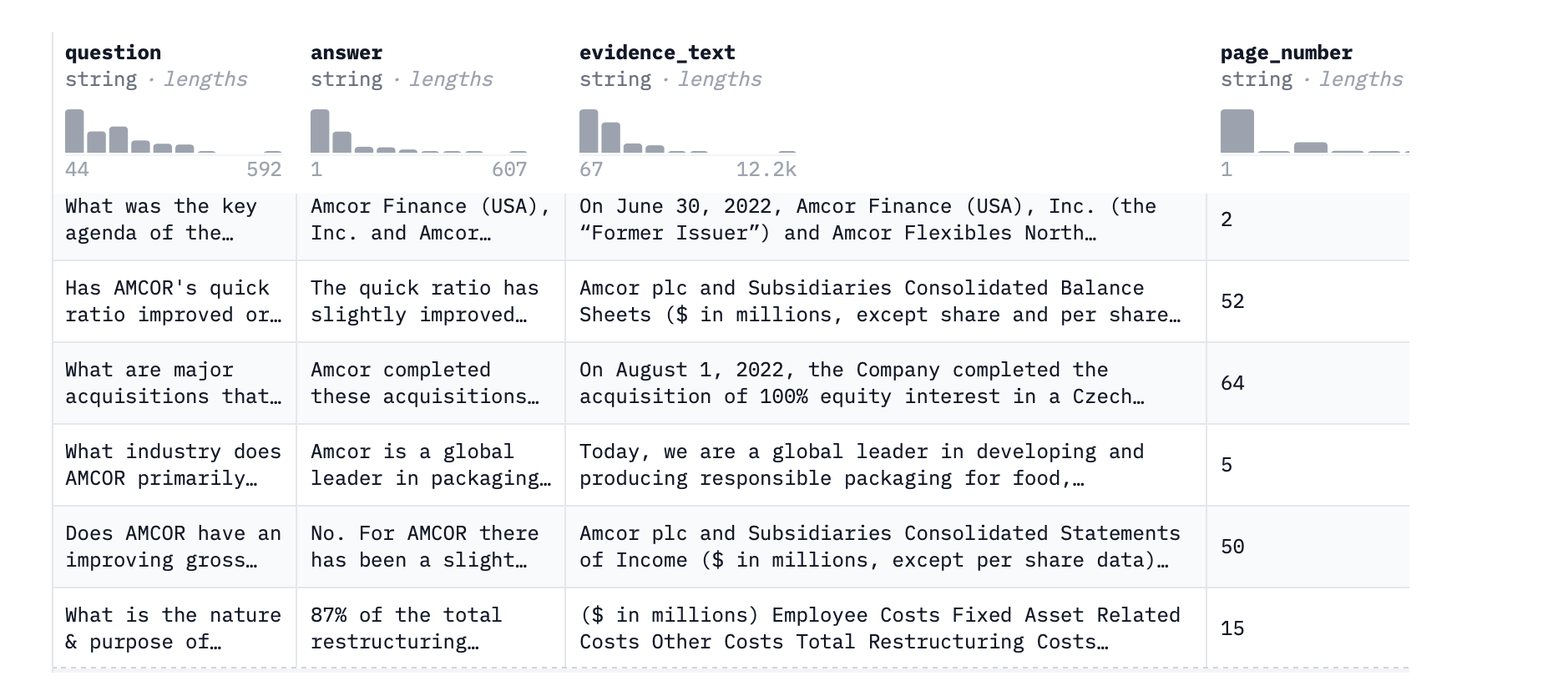}
\end{figure} 

\footnotetext{Finance Bench Dataset. \url{https://huggingface.co/datasets/PatronusAI/financebench}, Accessed on 2023-01-24}

\begin{figure}[h!]
\centering
\caption{Sample of RAG Instruct Benchmark Tester Dataset \protect\footnotemark}
\includegraphics[scale=0.45]{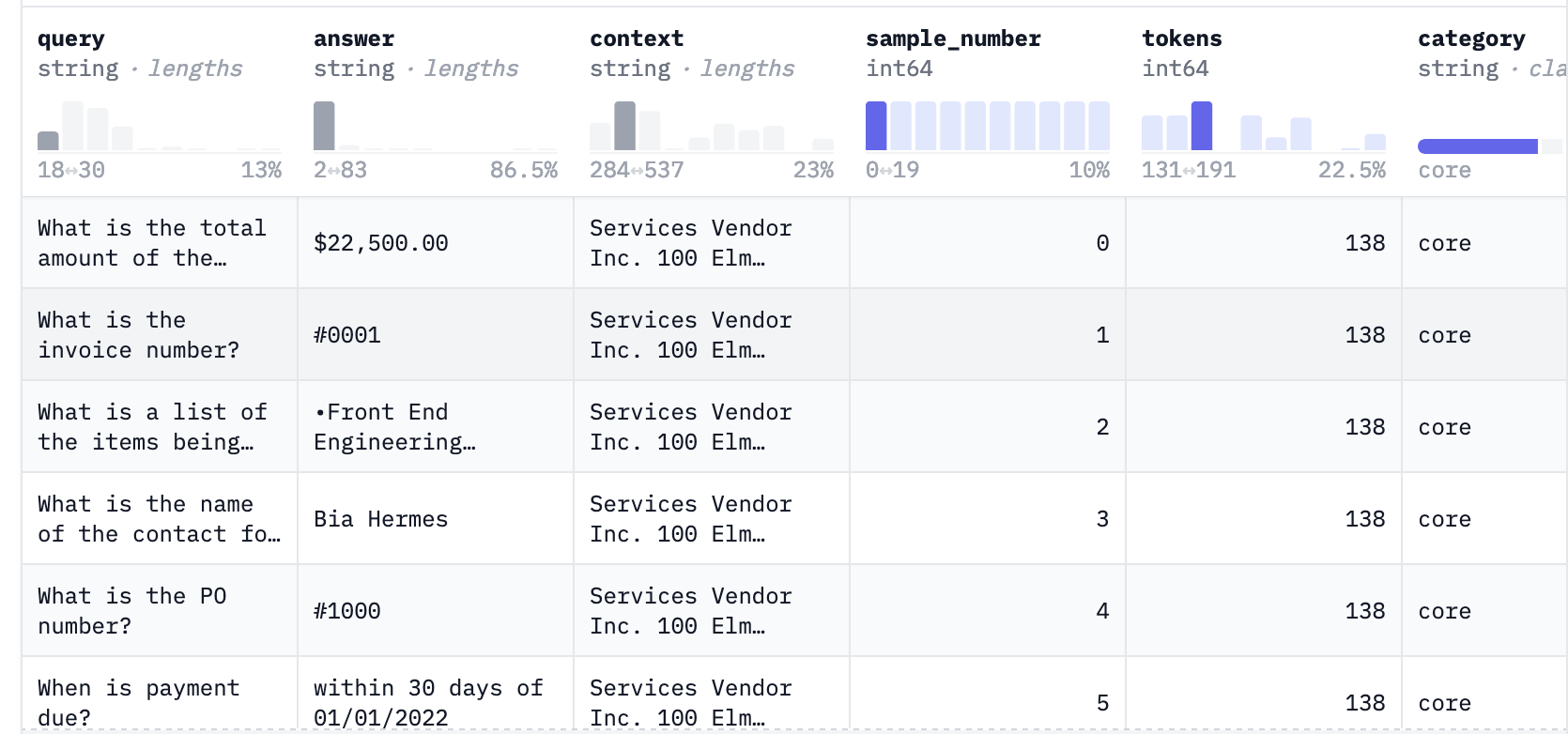}
\end{figure} 

\footnotetext{rag\_instruct\_benchmark\_tester Dataset. \url{https://huggingface.co/datasets/llmware/rag_instruct_benchmark_tester}, Accessed on 2023-01-24}

\subsection{Data Preprocessing}

The data had to be formatted in a very specific way for fine-tuning each large language model. Both open-source LLMs like LLaMA-2 and closed LLMs like GPT-3.5 have a certain format that the prompt and data must be presented based on the way the model was developed. This process involved concatenating the different columns from the data sets in a specific way as well as incorporating any sort of prompt engineering. 
\\
\\
For example, the LLaMA-2 model requires data in the following format: 
\vspace{1em}

\begin{lstlisting}[language=Python, breaklines=true]
<s>[INST] <<SYS>>\n{system_prompt}\n<</SYS>>\n\n{user_prompt} {input_section} [/INST] {model_answer} </s>
\end{lstlisting}

\vspace{1em}

\noindent
The meta tags specify user vs. system. Instructions give the model a lot more direction and yield much better results than just arbitrarily concatenating the different columns. Within the system text, we can utilize prompt engineering techniques to assign the LLM a specific role that will make it behave in a certain way. For example, when fine-tuning LLaMA-2 on the FinanceBench Dataset, the following prompt was used: 

\vspace{1em}

\noindent \textit{"You are a financial chatbot trained to answer questions based on the information provided in 10-K documents. Your responses should be directly sourced from the content of these documents. When asked a question, ensure that your answer is explicitly supported by the text in the 10-K filing, and do not include any external information, interpretations, or assumptions not clearly stated in the document. If a question pertains to financial data or analysis that is not explicitly covered in the 10-K filing provided, respond by stating that the information is not available in the document. Your primary focus should be on accuracy, specificity, and adherence to the information in 10-K documents, particularly regarding financial statements, company performance, and market positions"}
\vspace{1em}

\noindent
This is an example of how prompt engineering can help reduce hallucination as it is very clearly specified that the model should not make up information if the answer is not found in the provided context. For this use case, we prioritize accuracy over creativity based on the context of how a model like this will be used. Experimenting with prompt engineering within the fine-tuning data provides another method to optimize these models for specific use cases. 
\\
\\
The user prompt is what is said to the model by the user and consists of both the question itself and the provided context. All of that is surrounded by instruction brackets to indicate to the model that it was all input, and based on said instructions, it should provide an output. In the training data, the answer is concatenated after the instruction tag is closed. For model inference, all the input text is fed into the model the same way as it was formatted in the training data, where the model is given everything within the [INST] brackets, and infers the response.

\section{Approach}
In order to enhance the existing LLMs, we must improve both the models' abilities to answer questions from a given context and their abilities to retrieve the most relevant context. The specific use case we have in mind is the ability to answer questions on financial documents, so the model must know how to answer finance-related questions and where in the documents to find the correct answers. We will go over our approach to improve both of these aspects through human feedback and other techniques 

\subsection{Enhancing Question-Answering}

The datasets we worked with had various questions, answers, and contexts curated by human experts. Each row with all this information can be thought of as a label, and we wanted to see how well the model improves after incrementally increasing the number of labels it is given. For the FinanceBench Dataset, we utilized this method for 4 LLMs: Claude, GPT 3.5 Turbo, GPT4ALL, and LLaMA-2. For Claude and GPT4ALL, the model weights were not available to be fine-tuned, so we included the labels via few-shot prompting. This means that all the human-labeled examples from the dataset were included in the query itself, but none of the model weights were actually modified. GPT 3.5 Turbo was fine-tuned via the OpenAI API on training data of varying lengths in the proper jsonl format required by the docs. Finally, LLaMA-2 is an open-source model available on the HuggingFace Hub (after authorization), and was fine-tuned using the HuggingFace Transformer's Trainer class. We incremented the number of training labels, with each increment by 10m to a total of 50 labels and conducted model inference on all fixed test rows of the FinanceBench Dataset. 
\\
\\
Essentially, for each LLM we developed 5 different versions with 10, 20, 30, 40 and 50 labels included as training data respectively. The training process for each version was the same, the only difference being the number of human-curated labels given. However, the fine-tuning or few show prompting processes for each model differed from each other in order to correspond to the documentation for each LLM. GPT 3.5 Turbo and LLaMA-2 were the only models that allowed fine-tuning that alters model parameters. For each of those models, a system prompt was included that specified the role of the LLM to be a financial chatbot that does not hallucinate information that is not included in the provided context. 
\\
\\
This method of fine-tuning only enhanced the model's ability to answer questions accurately given that it had access to the sections in the document that contained the answer. This is because the context was included in the given labels and concatenated to the prompt accordingly. The context was included and given to the LLM the same way it would be in a traditional RAG architecture, the only difference is that the context was given based on human expertise instead of retrieved by an algorithm. Therefore, to develop an optimized finance LLM, we need to develop a strong RAG model too, or our fine-tuned model will not be able to infer the answer from an incorrect chunk from the document.

\subsection{Enhancing Retrieval}
Although fine-tuning methods can enhance a model's ability to answer questions when provided with context, the overall performance will still suffer significantly if the model cannot retrieve the relevant chunks that contain the answer in the first place. In a traditional RAG system, the documents are first divided into chunks and converted into embeddings, which are numerical representations of the text. The most relevant chunks are found by conducting a cosine similarity search between the query and all the embeddings. The chunks with the most similar embeddings, measured through a cosine similarity score, are what gets returned by the retrieval function. The LLM uses these sections of the text as context to answer the question. 
\\
\\
However, the simple RAG pipeline often fails for more complicated documents and advanced question-answering. This is because the sections of the document that most likely contain the answers, or the sections a human expert would look to are not necessarily the sections with the most similar words. The reality is much more nuanced than that as most similar does not necessarily mean most relevant. The challenge is then to figure out how to develop an algorithm that can account for such nuances. In this paper, we employed a technique called Forward-looking active retrieval augmented generation (FLARE) to try and address the pitfalls of traditional RAG techniques. 
\\
\\
FLARE employs a much more active approach rather than a one-and-done retrieval pipeline. Instead of doing a similarity search of just the query and document embeddings, FLARE prompts the LLM to generate a hypothetical response from the query without context, and then both the query and hypothetical response are used to trigger the RAG step to search for the most similar chunks.\footnote {FLARE — “Advanced” RAG implemented from scratch \url {https://ayushtues.medium.com/flare-advanced-rag-implemented-from-scratch-07ca75c89800} Accessed on 2024-01-18} This aspect is known as Hypothetical Document Embeddings(HyDE) and is a version of query expansion. FLARE builds on top of this by setting a threshold when to and when to not trigger retrieval. Only if the next generated tokens have a probability under a certain threshold, meaning that the model is not confident in the next predictions, it will go and try to retrieve relevant chunks from the document. This way, the model does not have to rely on the initial chunks it retrieved but can actively update the context based on how confident it is.

\section{Evaluating Method}  
In assessing how well each LLM performs, we used a detailed method with three measures: ROUGE-L, cosine similarity, and LLM evaluation. We looked at datasets with questions, answers, and context made by human experts to see how well the models understand and respond. ROUGE-L checked how closely the model's responses matched human answers, while cosine similarity checked how well the model's outputs aligned with human references. For LLM evaluation, we prompted GPT 3.5 to give a score of 0 or 1 based on whether the model response and the reference answer conveyed the same idea. We tested six versions of each LLM, starting with 0 labeled data points and going up to 50, to see how more labeled data affected performance. For fine-tuning, each model had its own method, including few-shot prompting, OpenAI API-based fine-tuning, or using the HuggingFace Transformer’s Trainer class. Some models, like GPT 3.5 Turbo and LLaMA-2, were fine-tuned with specific prompts to act as financial chatbots. This thorough evaluation, considering both language and meaning, showed how adding labeled data, from 0 to 50, impacted each LM's performance. We also conducted benchmarking on the Flare RAG dataset using a vector database, employing a similar approach.

\subsection{Choosing the Right Approach}
There are various models available and different methods to improve each one for your specific use case, each having its own trade offs. A key point is the difference between open-source and closed-source LLMs and the implications it has with respect to privacy. Many firms have prohibited the use of ChatGPT because of concerns of data privacy as they might want to use these tools to ask questions and interact with confidential documents. If data privacy is a primary concern, then Open source models like LLAMA-2 are the only options as they can be downloaded locally or hosted in a cloud based system that one has control over. However, the primary trade off of these private models is that it requires having access to more compute, usually several GPUs to have comparable speed to closed source models, and this can lead to significant additional costs. If data privacy is not as big of a concern, GPT and Claude are great options, yet one needs to be mindful of the cost of API calls and how that scales depending on the use case. Most open source models are significantly smaller in size, in this paper we used the 7 billion parameter version of LLAMA, while models like GPT have around 175 billion parameters. However, through proper fine tuning and implementation of other techniques, a smaller model can be optimized for a specific use case to be comparable to the larger models. Ultimately, it is important to consider both your priorities and resources available to choose the best method. 

\begin{figure}[h!]
\centering
\caption{Besides Accuracy, Other Factors to Consider When Choosing your LLM \protect\footnotemark}
\includegraphics[scale=0.45]{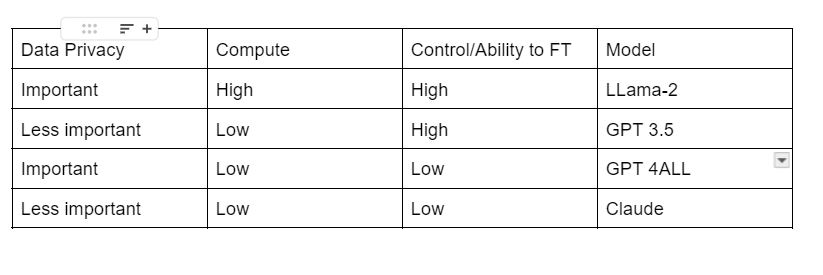}
\end{figure} 

\section{Results}
In the plots below, you can see the rouge score and the cosine similarity score increases more or less when the number of labels increases, demonstrating the effectiveness of fine tuning. To see the raw
results and code in more detail, please access it \href{https://github.com/nv78/Anote/tree/main/Benchmarking-Question-Answering}{here}.

\begin{figure}[h!]
\centering
\caption{Finance Bench Dataset}
\subcaption[]{Claude Results}
\includegraphics[scale=0.37]{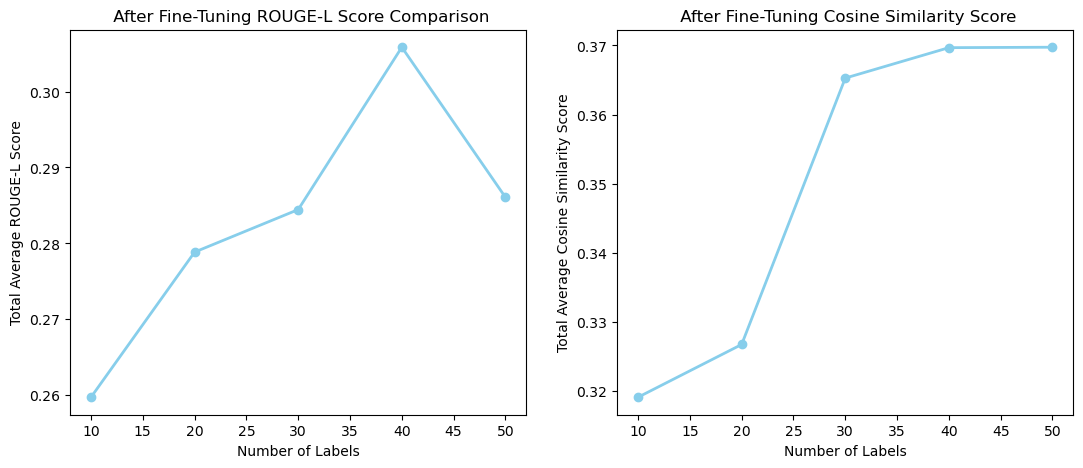}
\subcaption[]{Fine tuned GPT 3.5 Results}
\includegraphics[scale=0.37]{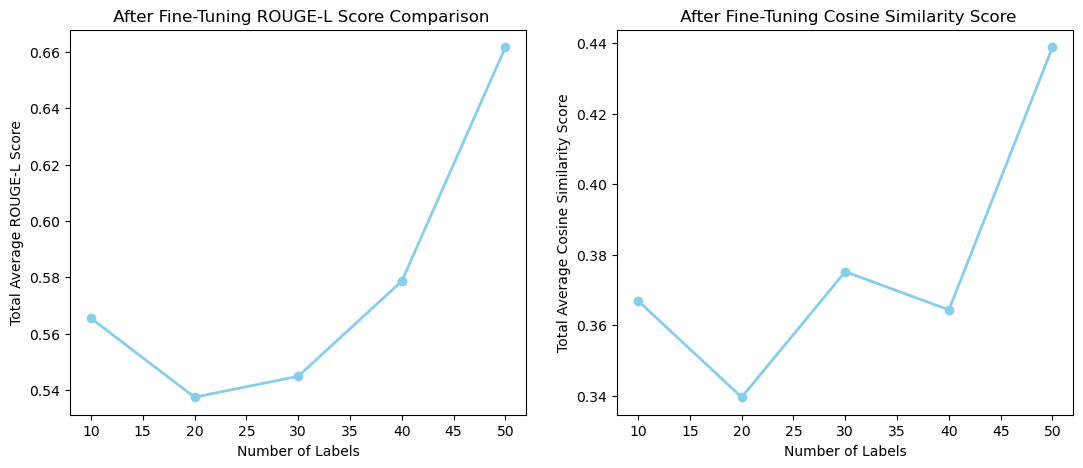}
\subcaption[]{Llama2 Results}
\includegraphics[scale=0.37]{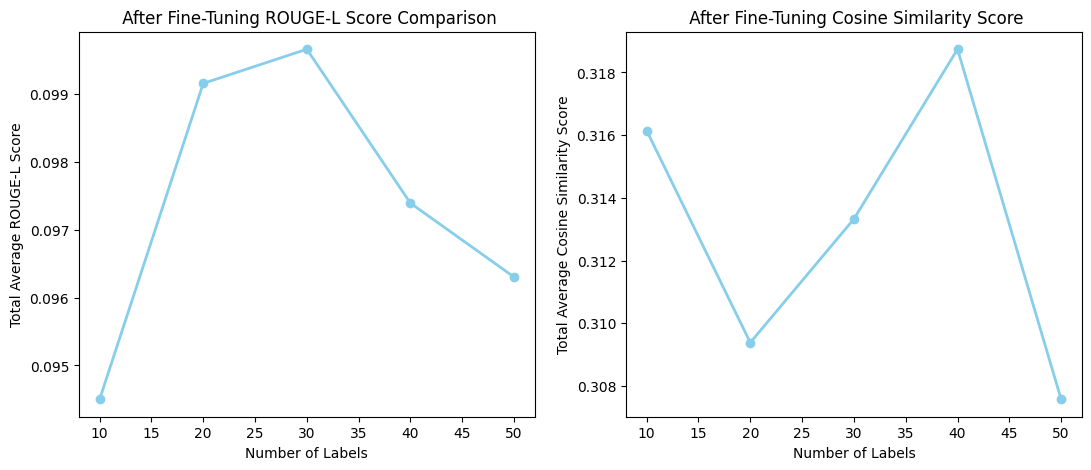}
\subcaption[]{GPT4ALL Results}
\includegraphics[scale=0.37]{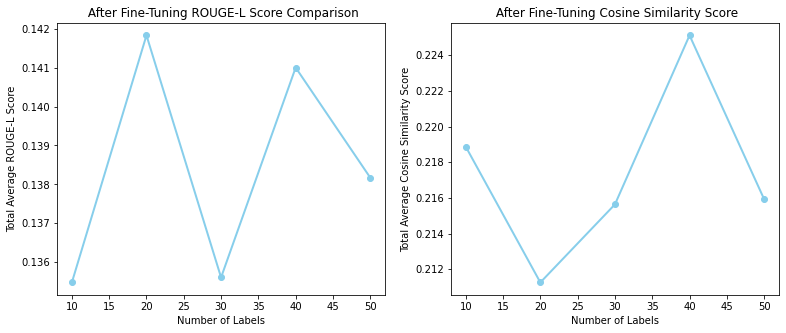}
\subcaption[]{FLARE Results}
\includegraphics[scale=0.37]{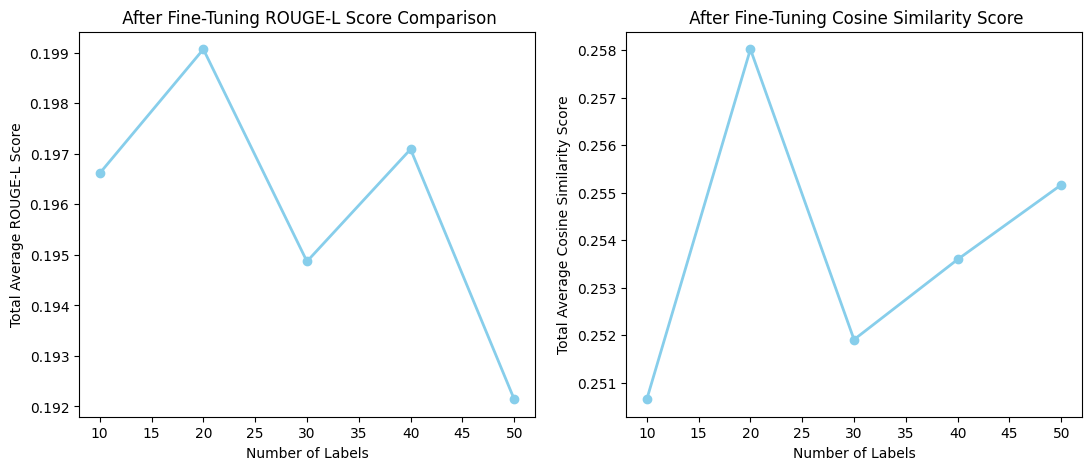}
\end{figure} 

\begin{figure}[h!]
\centering
\caption{Rag Instruct Dataset}
\subcaption[]{Claude Results}
\includegraphics[scale=0.45]{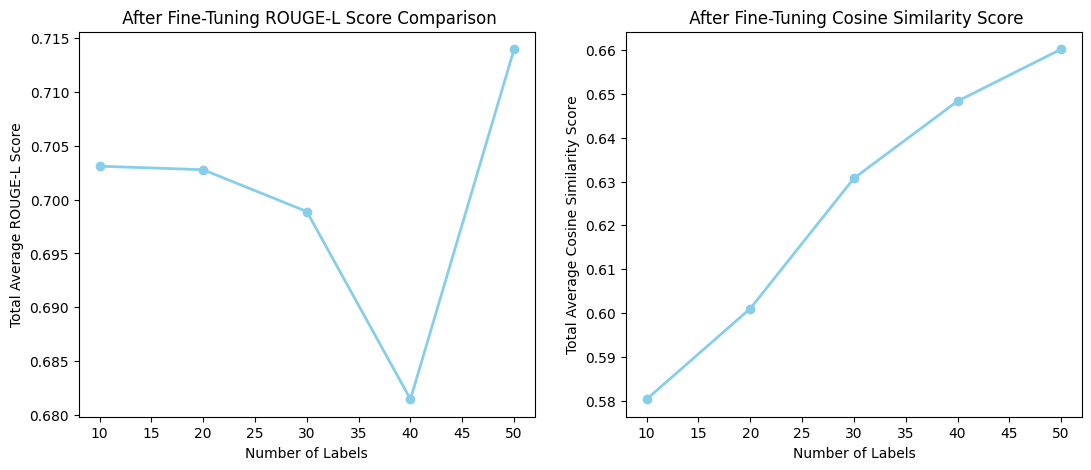}
\subcaption[]{Fine Tuned GPT 3.5 Results}
\includegraphics[scale=0.45]{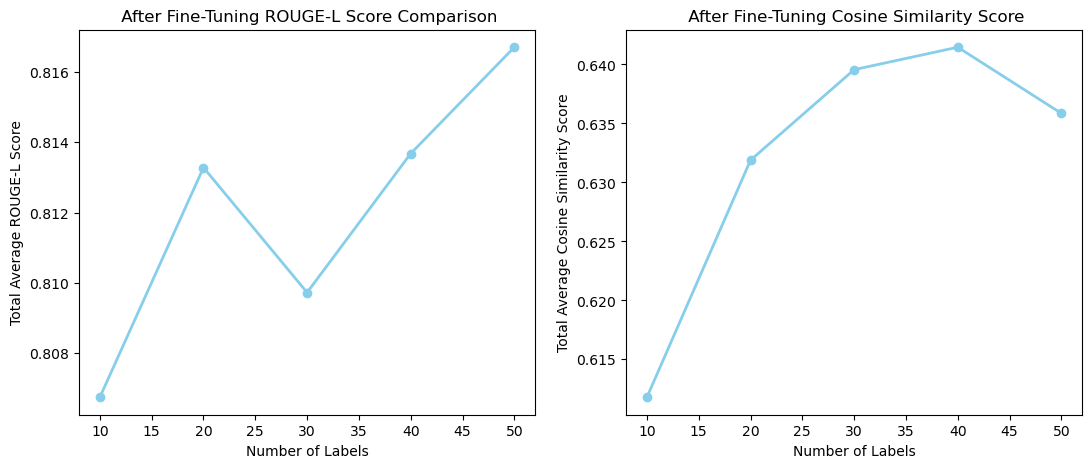}
\subcaption[]{GPT4ALL Results}
\includegraphics[scale=0.45]{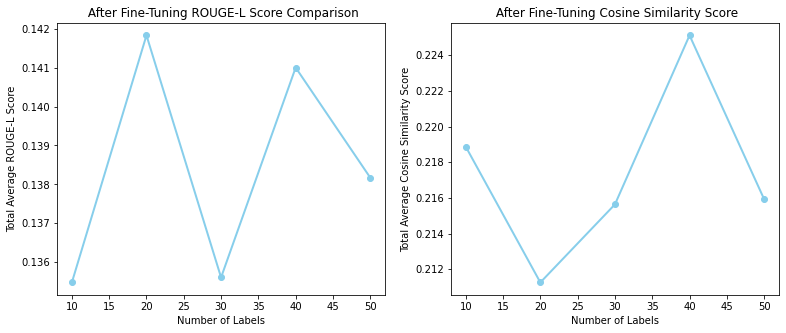}
\subcaption[]{FLARE Results}
\includegraphics[scale=0.45]{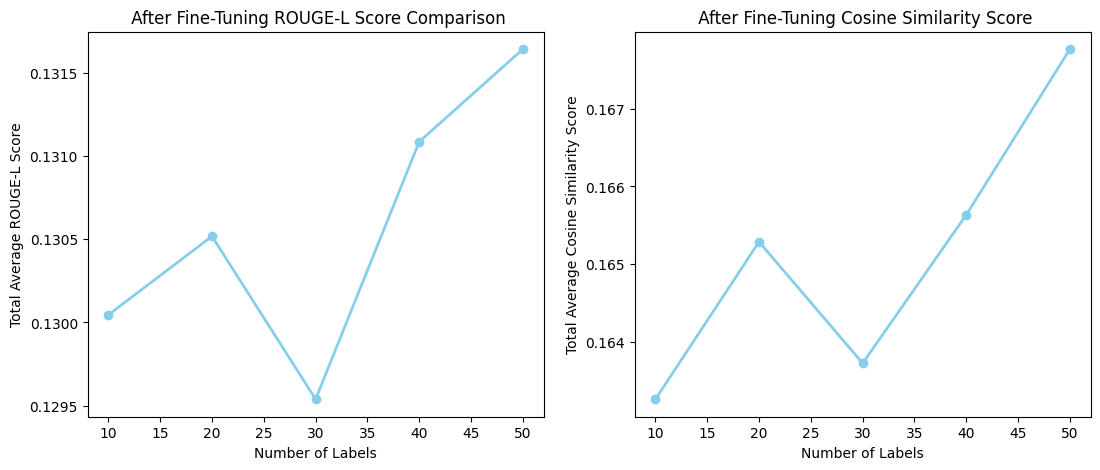}
\end{figure}

\clearpage
\section{Conclusion}
In summary, when we fine-tuned the models by incorporating feedback from humans and adding more labeled examples, we consistently saw improved accuracy across various datasets. Training the model by adding a few labeled samples in edge cases or specific areas where the model struggled made a big difference. By focusing on these weak points through repeated fine-tuning and adjustment of prompts, the model got better over time. This iterative process, guided by user feedback, played a crucial role in boosting the overall accuracy of the model. \\
\\
\noindent Another key finding was that simple RAG pipelines do not perform well on domain specific questions and documents, so other techniques need to be employed to develop a more accurate retrieval algorithm. A robust document based question-answering system is contingent on having both a highly accurate and specialized LLM as well as a robust retrieval method that provides the most relevant context. To check out the detailed results and code, you can find them by accessing the provided link.

\section{Next Steps} 

For next steps, one thing we can do is more testing to tune different model parameters to see if it would enhance performance. Only chunk size and chunk overlap were adjusted in RAG, but most of the training parameters were kept at the default for the training process. Testing out different embedding models, or even fine tuning the embedding model for the finance specific domain would be a method to try in the near future. Additionally, while supervised fine tuning (SFT) was the primary technique used in this paper, it is also possible to in corporate other methods such as unsupervised fine tuning and reinforcement learning with human feedback into these models to further improve performance. We are also looking to test out more methods to improve retreival algorithms in addition to FLARE and HyDE to help find the best and optimal chunk of text. These include implementing a re-ranking algorithm, such as Cohere's that will rearrange the document chunks based on relevant instead of just similarity.\footnote {Nils Reimers, Sylvie Shi, Lucas Fayoux, Elliott Choi, Say Goodbye to Irrelevant Search Results: Cohere Rerank Is Here, May 2023 \url{https://txt.cohere.com/rerank/}} Since we are working with a variety of financial documents with various different and complex sections, metadata annotations could provide some value to further filter document chunks for RAG. In summary, we will continue to strive for more ways to make these models and the fine tuning process faster and more efficient while maintaining and improving model performance.

\section{References}


\small
\begin{enumerate}
    
    \item Patrick Lewis, Ethan Perez, Aleksandra Piktus, Fabio Petroni, Vladimir Karpukhin, Naman Goyal, Heinrich Küttler, Mike Lewis, Wen-tau Yih, Tim Rocktäschel, Sebastian Riedel, Douwe Kiela. Retrieval-Augmented Generation for Knowledge-Intensive NLP Tasks, 2020.

    \item Disadvantages of RAG. \url{https://medium.com/@kelvin.lu.au/disadvantages-of-rag-5024692f2c53}, Accessed on 2024-01-22.

    \item Pranab Islam1, Anand Kannappan, Douwe Kiela,
        Rebecca Qian, Nino Scherrer, Bertie Vidgen. FINANCEBENCH: A New Benchmark for Financial Question Answering, 2023. arXiv:2311.11944.

    \item Edward Hu, Yelong Shen, Phillip Wallis, Zeyuan Allen-Zhu,
    Yuanzhi Li, Shean Wang, Lu Wang, Weizhu Chen. LORA: LOW-RANK ADAPTATION OF LARGE LANGUAGE MODELS, 2021. arXiv:2106.09685.

    \item Ayush Mangal. FLARE — “Advanced” RAG implemented from scratch, 2023. \url{https://ayushtues.medium.com/flare-advanced-rag-implemented-from-scratch-07ca75c89800}

    \item Quantized Fine-tuning on LLaMa-2 for RAG, 2023. \url {https://medium.com/towards-generative-ai/quantized-fine-tuning-llama-2-for-rag-question-answering-426114daa09a}

    \item Maxime Labonne. Fine-Tune Your Own Llama 2 Model in a Colab Notebook, 2023. \url{https://towardsdatascience.com/fine-tune-your-own-llama-2-model-in-a-colab-notebook-df9823a04a32}

    \item Philipp Schmid, Omar Sanseviero, Pedro Cuenca, Lewis Tunstall. Llama 2 is here - get it on Hugging Face, 2023. \url{https://huggingface.co/blog/llama2}

    \item How to Fine-tune Llama 2 with LoRA for Question Answering: A Guide for Practitioners, 2023. \url{https://deci.ai/blog/fine-tune-llama-2-with-lora-for-question-answering/}

    \item Hugo Touvron, Thibaut Lavril, Gautier Izacard, Xavier Martinet, Marie-Anne Lachaux, Timothee Lacroix, Baptiste Rozière, Naman Goyal, Eric Hambro, Faisal Azhar, Aurelien Rodriguez, Armand Joulin, Edouard Grave, Guillaume Lample. LLaMA: Open and Efficient Foundation Language Models, 2023. arXiv:2302.13971. 

    \item Leonie Monigatti. A Guide on 12 Tuning Strategies for Production-Ready RAG Applications, 2023. \url{https://towardsdatascience.com/a-guide-on-12-tuning-strategies-for-production-ready-rag-applications-7ca646833439}

    \item Matt Ambrogi. 10 Ways to Improve the Performance of Retrieval Augmented Generation Systems, 2023. \url{https://towardsdatascience.com/10-ways-to-improve-the-performance-of-retrieval-augmented-generation-systems-5fa2cee7cd5c}

    \item Nathan Lambert, Louis Castricato, Leandro von Werra, Alex Havrilla. Illustrating Reinforcement Learning from Human Feedback (RLHF), 2022. \url{https://huggingface.co/blog/rlhf}

    \item Luyu Gao, Xueguang Ma, Jimmy Lin, Jamie Callan. Precise Zero-Shot Dense Retrieval without Relevance Labels, 2022. arXiv:2212.10496. 

    \item Sourab Mangrulkar, Sayak Paul. PEFT: Parameter-Efficient Fine-Tuning of Billion-Scale Models on Low-Resource Hardware, 2023. \url{https://huggingface.co/blog/peft}

    \item Tim Dettmers, Artidoro Pagnoni, Ari Holtzman, Luke Zettlemoyer. QLORA: Efficient Finetuning of Quantized LLMs, 2023. arXiv:2305.14314.

\end{enumerate}

\section{Appendix A: Dataset Descriptions}

\begin{enumerate}
\item Finance Bench Dataset
\begin{itemize}
    \item Description: Question and answer dataset for 10-K documents
    \item Source: PatronusAI on HuggingFace 
    \item Size: 150 rows
    \item URL: \url{https://huggingface.co/datasets/PatronusAI/financebench}
\end{itemize}
\item RAG Instruct Benchmark Tester Dataset
\begin{itemize}
    \item Description: Question and answer dataset for various financial documents
    \item Source: LLMWare on HuggingFace 
    \item Size: 200 rows
    \item URL: \url{https://huggingface.co/datasets/llmware/rag_instruct_benchmark_tester}
\end{itemize}
\end{enumerate}

\end{document}